\RequirePackage{fix-cm}
\documentclass[twocolumn]{svjour3}          
\pdfoutput=1
\smartqed  
\usepackage{graphicx}
\usepackage{biblatex}
\usepackage{multirow}
\usepackage{amsmath}
\usepackage{amssymb}
\usepackage{bbm}
\usepackage{bm}
\usepackage{booktabs}
\usepackage{array}
\usepackage{blindtext}
\usepackage{comment}

\usepackage{pifont}
\newcommand{\cmark}{\ding{51}}

\usepackage{xcolor}
\usepackage{color, colortbl}
\definecolor{citecolor}{HTML}{0071bc}
\definecolor{tabhighlight}{HTML}{e5e5e5}
\usepackage {hyperref} 
\hypersetup {colorlinks,citecolor=citecolor}

\usepackage{wrapfig,caption,subcaption}
\usepackage{tabulary,xspace,makecell}
\usepackage{dsfont,fixmath,mathtools,nicefrac}

\usepackage{color, colortbl}
\definecolor{deemph}{gray}{0.6}
\newcommand{\gc}[1]{\textcolor{deemph}{#1}}
\definecolor{GrayBG}{gray}{0.95}
\definecolor{Highlight}{HTML}{39b54a}  
\definecolor{Green}{HTML}{39b54a}
\definecolor{Purple}{RGB}{167,1,233}
\newcommand{\GrayBG}[1]{\cellcolor{GrayBG}{#1}}

\providecommand{\eg}{\textit{e.g.}\@\xspace}
\providecommand{\ie}{\textit{i.e.}\@\xspace}
\newcommand{\gxmark}{\textcolor{black!20}{\ding{55}}}

\newcommand{\fixedap}{AP$^\text{Fix}$\xspace}
\newcommand{\fixedapr}{AP$^\text{Fix}_\textrm{r}$\xspace}
\newcommand{\fixedapc}{AP$^\text{Fix}_\textrm{c}$\xspace}
\newcommand{\fixedapf}{AP$^\text{Fix}_\textrm{f}$\xspace}
\newcommand{\apr}{AP$_\textrm{r}$\xspace}
\newcommand{\apc}{AP$_\textrm{c}$\xspace}
\newcommand{\apf}{AP$_\textrm{f}$\xspace}
\newcommand{\apone}{AP$_\textrm{1}$\xspace}
\newcommand{\aptwo}{AP$_\textrm{2}$\xspace}
\newcommand{\apthree}{AP$_\textrm{3}$\xspace}
\newcommand{\apfour}{AP$_\textrm{4}$\xspace}

\definecolor{demphcolor}{RGB}{90,90,90}
\newcommand{\demph}[1]{\textcolor{Highlight}{#1}}
\newcommand{\dt}[1]{\fontsize{6pt}{0.1em}\selectfont \demph{(#1)}}
\newcommand{\tablestyle}[2]{\setlength{\tabcolsep}{#1}\renewcommand{\arraystretch}{#2}\centering\footnotesize}
\newcolumntype{x}[1]{>{\centering\arraybackslash}p{#1pt}}
\newlength\savewidth
\newcommand{\td}[1]{\scriptsize\rlap{ \dt{#1}}}
\newcommand{\smallsec}[1]{\vspace{0.5ex}\noindent\textbf{#1}\quad}
\DeclareUnicodeCharacter{2212}{-}

\makeatletter
\renewcommand\paragraph{
  \@startsection{paragraph} 
  {4} 
  {\z@} 
  {.5em \@plus1ex \@minus.2ex} 
  {-.5em} 
  {\normalfont\normalsize\bfseries} 
}
\makeatother
\journalname{IJCV}
\begin{document}
\sloppy

\title{Semi-Supervised and Long-Tailed Object Detection \\ with CascadeMatch}


\author{Yuhang Zang \and
        Kaiyang Zhou \and
        Chen Huang \and
        Chen Change Loy
}


\institute{Yuhang Zang \at
              S-Lab, Nanyang Technological University, Singapore \\
              \email{zang0012@ntu.edu.sg}
           \and
           Kaiyang Zhou \at
              S-Lab, Nanyang Technological University, Singapore \\
              \email{kaiyang.zhou@ntu.edu.sg}
           \and
           Chen Huang \at
              Apple Inc., USA \\
              \email{chen-huang@apple.com}
           \and
           Chen Change Loy (corresponding author) \at
              S-Lab, Nanyang Technological University, Singapore \\
              \email{ccloy@ntu.edu.sg}
}

\date{Received: date / Accepted: date}

\maketitle

\begin{abstract}
This paper focuses on long-tailed object detection in the semi-supervised learning setting, which poses realistic challenges, but has rarely been studied in the literature. 
We propose a novel pseudo-labeling-based detector called CascadeMatch. Our detector features a cascade network architecture, which has multi-stage detection heads with progressive confidence thresholds. To avoid manually tuning the thresholds, we design a new adaptive pseudo-label mining mechanism to automatically identify suitable values from data.
To mitigate confirmation bias, where a model is negatively reinforced by incorrect pseudo-labels produced by itself, each detection head is trained by the ensemble pseudo-labels of all detection heads. Experiments on two long-tailed datasets, i.e., LVIS and COCO-LT, demonstrate that CascadeMatch surpasses existing state-of-the-art semi-supervised approaches---across a wide range of detection architectures---in handling long-tailed object detection. For instance, CascadeMatch outperforms Unbiased Teacher by 1.9 \fixedap on LVIS when using a ResNet50-based Cascade R-CNN structure, and by 1.7 \fixedap when using Sparse R-CNN with a Transformer encoder. We also show that CascadeMatch can even handle the challenging sparsely annotated object detection problem. Code: \href{https://github.com/yuhangzang/CascadeMatch}{https://github.com/yuhangzang/CascadeMatch}.
\end{abstract}

\section{Introduction} \label{sec:intro}
\begin{figure*}[!htb]
\begin{minipage}{0.32\textwidth}
\centering
\includegraphics[width=\linewidth]{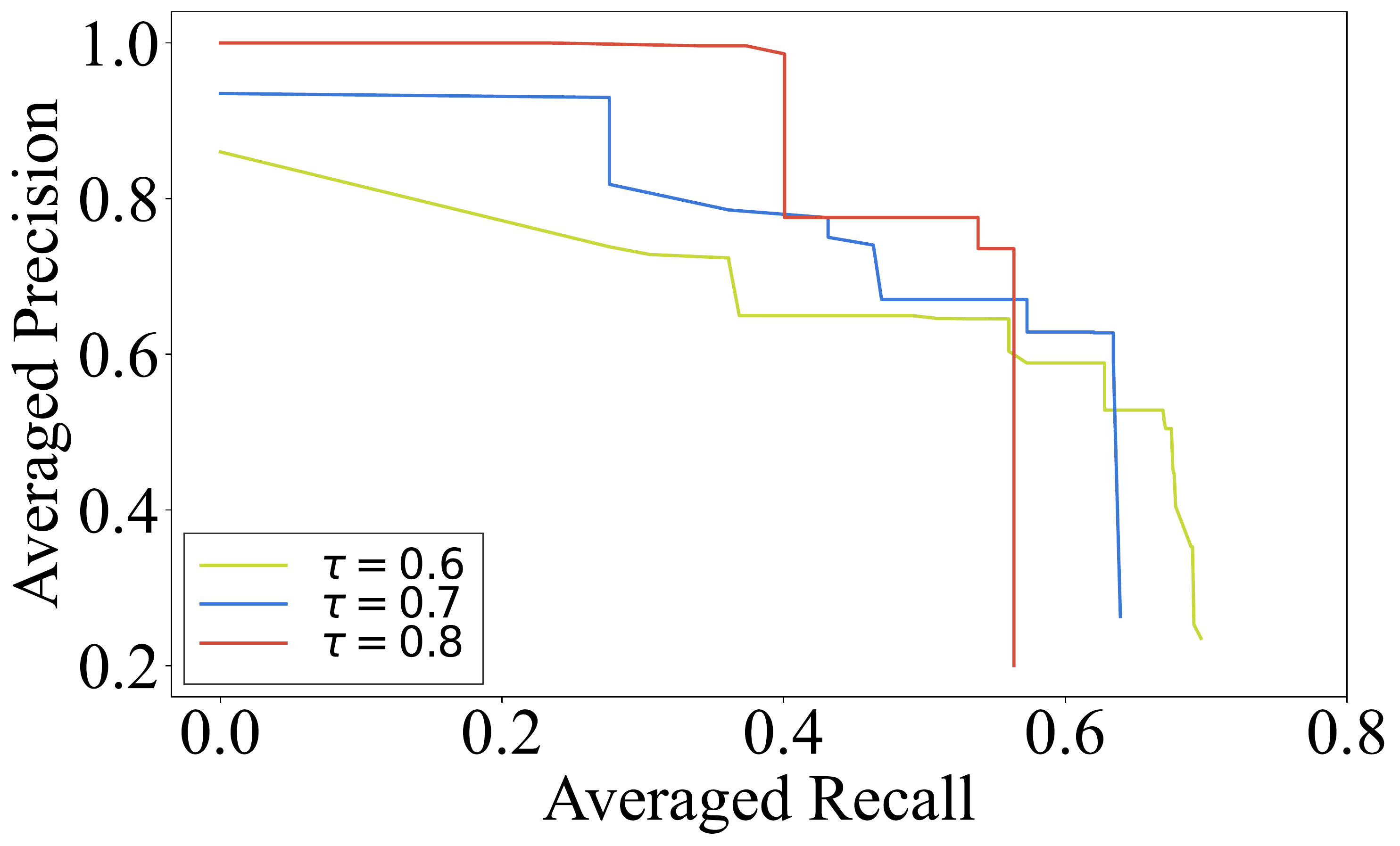}
\end{minipage}
\hfill
\begin{minipage}{0.32\textwidth}
\centering
\includegraphics[width=\linewidth]{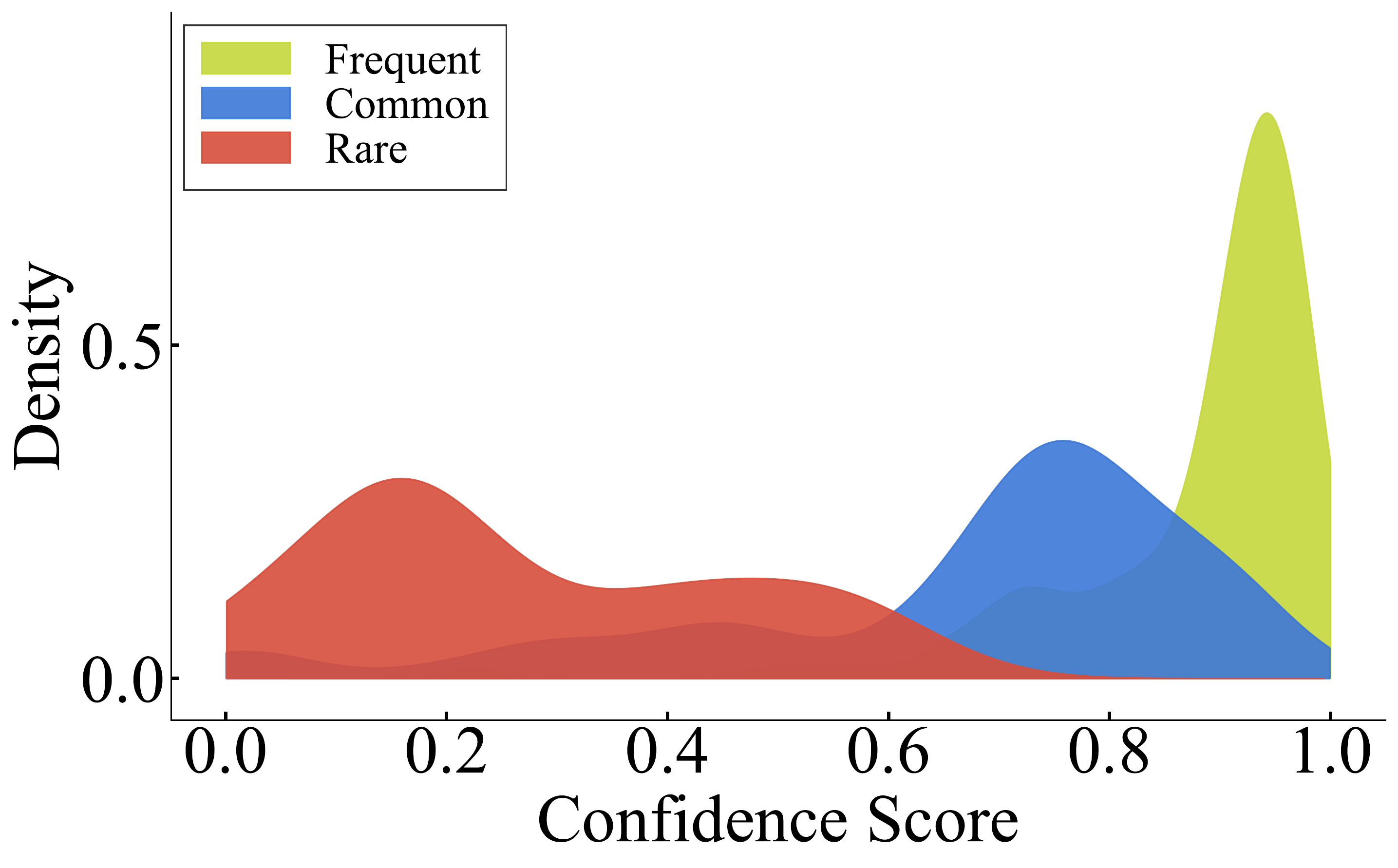}
\end{minipage}
\hfill
\begin{minipage}{0.32\textwidth}
\centering
\includegraphics[width=\linewidth]{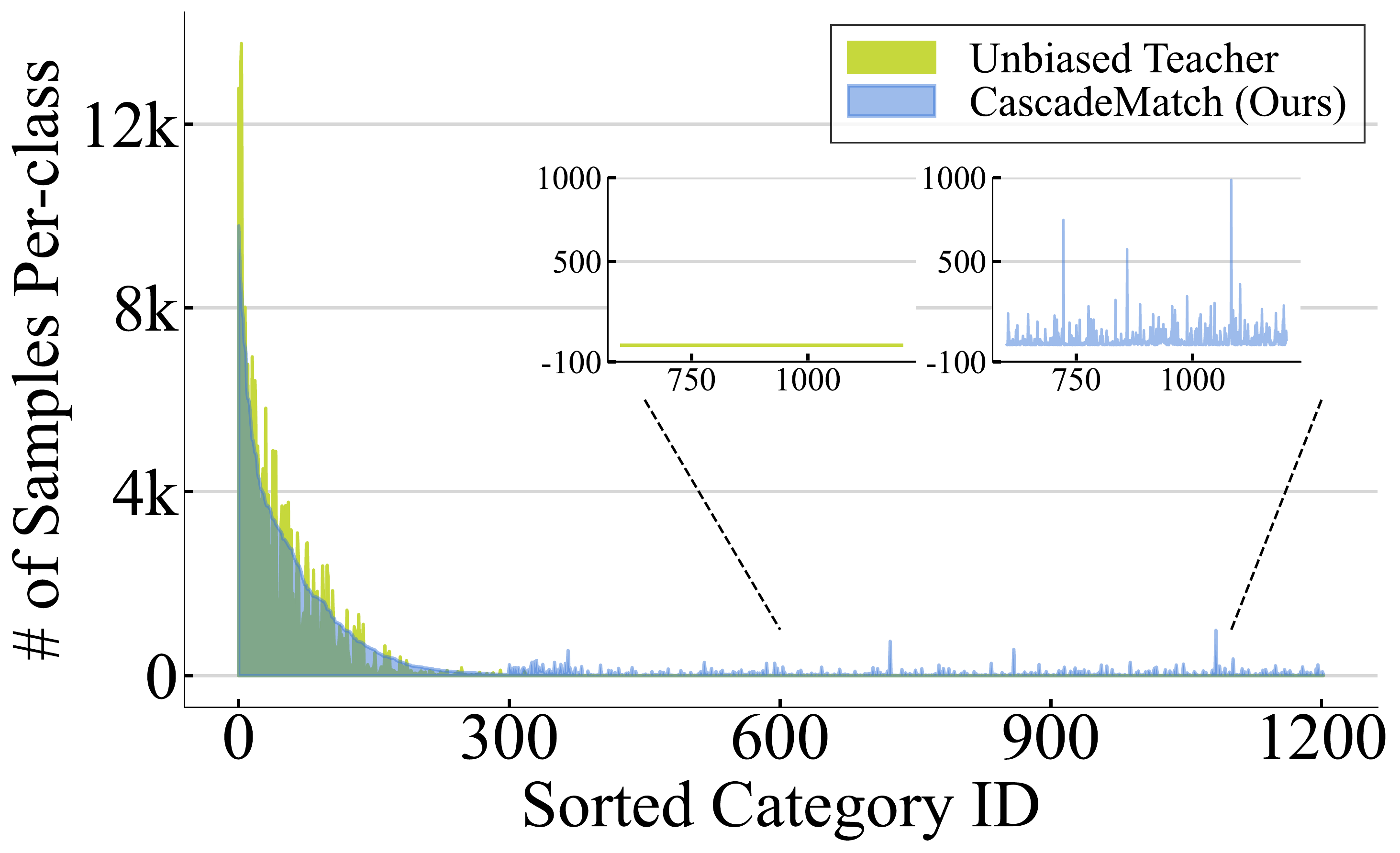}
\end{minipage}
\caption{
Motivation of our research.
\textbf{(a)} The Average Precision (AP) and Average Recall (AR) curves, obtained using different \emph{fixed} confidence thresholds (denoted by $\tau$). Clearly, none of the chosen thresholds gives the best trade-off.
\textbf{(b)} The distribution of prediction scores for a long-tailed dataset, which shows a high degree of imbalance between the three class groups.
\textbf{(c)} Sorted number of samples per class seen by the model during training. CascadeMatch retains much more pseudo-labeled samples than Unbiased Teacher with respect to the common and rare classes.
}
\label{fig:introduction}
\end{figure*}

Though object detection has been significantly advanced in the supervised learning domain by neural network-based detectors~\cite{liu2016ssd,ren2015faster,lin2017focal,tian2019fcos,carion2020end}), there is still a large room for improvement in semi-supervised object detection (SSOD). In practice, SSOD is desirable because annotating bounding boxes and their object classes are both costly and time-consuming. Most existing semi-supervised object detectors~\cite{sohn2020simple,liu2021unbiased,zhou2021instant,tang2021proposal,tang2021humble,arazo2020pseudo,wang2021data,yang2021interactive}) are learned by estimated pseudo-labels, which are assigned to bounding box proposals and filtered by a single fixed confidence threshold. Such a combination of pseudo-labeling and confidence thresholds-based filtering has been largely inspired by research on semi-supervised image classification~\cite{berthelot2019mixmatch,xie2020unsupervised,sohn2020fixmatch,rizve2021defense}).

Most existing studies are conducted on the COCO dataset~\cite{lin2014microsoft} that has curated categories and highly balanced data distributions. However, real-world problems are much more challenging than what the COCO dataset represents in that data distributions are often \emph{long-tailed}, \ie, a majority of classes have only a few labeled images, which could easily result in an extremely biased detector. In recent years, the research community has paid increasing attention to long-tailed object detection, with several relevant datasets released, such as LVIS~\cite{gupta2019lvis} and COCO-LT~\cite{wang2020devil}. However, to our knowledge, \emph{none of the existing studies has been devoted to long-tailed object detection in the semi-supervised setting}, a more challenging yet practical problem.

Implementing semi-supervised object detection algorithms on long-tailed datasets is not trivial. By training a state-of-the-art semi-supervised detector, \ie, Unbiased Teacher~\cite{liu2021unbiased}, using a long-tailed LVIS~\cite{gupta2019lvis} dataset, we identify the following three major problems. First, a fixed confidence threshold often fails to provide a good trade-off between precision and recall. The shortcoming is evidenced in Figure~\ref{fig:introduction}(a), which shows none of the commonly used thresholds gives the best performance in both the AP and AR metrics, \eg, a fixed threshold of 0.6 returns the highest recall but has the lowest precision. Second, by digging deeper into the distribution of prediction scores, we observe that the model's predictions are biased toward the frequent classes (see Figure~\ref{fig:introduction}(b)). Finally, we identify the reason why using a fixed threshold leads to low confidence---and hence low prediction accuracy---on the common and rare classes: the model's exposure to these classes during training is substantially reduced compared to that to the frequent classes (see Figure~\ref{fig:introduction}(c)).

To overcome these problems, we propose \emph{CascadeMatch}, a novel pseudo-labeling-based approach to addressing long-tailed and semi-supervised object detection.
Specifically, CascadeMatch features a cascade pseudo-labeling~(CPL) design, which contains multi-stage detection heads.
To control the precision-recall trade-off, we set \emph{progressive} confidence thresholds for detection heads to focus on different parts.
The early detection head is assigned a small confidence threshold to improve recall, while the subsequent heads are assigned larger confidence thresholds to ensure precision.
The use of multiple heads also allows the unique chance for us to deal with confirmation bias -- a phenomenon where a model is iteratively reinforced by incorrect pseudo labels produced by itself. In particular, we show the possibility of using ensemble predictions from all detection heads as the teacher's supervision signal to obtain more reliable pseudo labels for training each individual detection head. 
To deal with the issue of biased prediction score distributions to frequent classes, we propose an adaptive pseudo-label mining mechanism~(APM) that automatically identifies suitable class-wise threshold values from data with minimal human intervention.
As shown in Figure~\ref{fig:introduction}(c), with the APM module, our approach can retain more pseudo-labels for common and rare classes than the previous SOTA approach~\cite{liu2021unbiased}, boosting the performance for classes with small sample sizes.

We present comprehensive experiments on two challenging long-tailed object detection datasets, namely LVIS v1.0~\cite{gupta2019lvis} and COCO-LT~\cite{wang2020devil}, \emph{under the SSOD setting}. Overall, CascadeMatch achieves the best performance on both datasets in all metrics. Notably, on LVIS, CascadeMatch improves upon the most competitive method, \ie, Unbiased Teacher~\cite{liu2021unbiased}, by 2.3\% and 1.8\% \fixedap in the rare and common classes, which confirm the effectiveness of our design for long-tailed data. Importantly, CascadeMatch is general and obtains consistent improvements \emph{across a variety of detection architectures}, covering both anchor-based R-CNN detectors~\cite{ren2015faster,cai2019cascade} and the recent Sparse R-CNN detector~\cite{sun2021sparse} with the Pyramid Vision Transformer encoder (PVT)~\cite{wang2021pyramid} (Table~\ref{tab:mainresults_lvis}). We also conduct various ablation studies to confirm the effectiveness of each of our proposed modules.

We also apply CascadeMatch to another challenging sparsely-annotated object detection (SAOD) setting~\cite{wu2018soft,zhang2020solving,wang2021co,zhou2021probabilistic} where training data are only partially annotated and contain missing annotated instances. Again, CascadeMatch yields considerable improvements over the supervised-only baseline and a state-of-the-art method~\cite{zhou2021probabilistic} (Table~\ref{tab:saod}). Finally, we provide several qualitative results and analyses to show that our proposed CascadeMatch method generates high-quality pseudo labels on both SSOD and SAOD settings.

\section{Related Work}

\smallsec{Semi-Supervised Object Detection}
has been a topical research area due to its importance to practical applications~\cite{rosenberg2005semi,misra2015watch,tang2016large,wang2018towards,roychowdhury2019automatic,jeong2019consistency,gao2019note,li2020improving,sohn2020simple,tang2021proposal,jeong2021interpolation,zhou2021instant,yang2021interactive,xu2021end,zhang2022semi,liu2022unbiased,chen2022dense,chen2022label,li2022semi,mi2022active,guo2022scale,li2022rethinking,liu2022unbiased}.
Various semi-supervised object detectors have been proposed in the literature, and many of them borrow ideas from the semi-supervised learning (SSL) community.
In CSD~\cite{jeong2019consistency} and ISD~\cite{jeong2021interpolation}, consistency regularization is applied to the mined bounding boxes for unlabeled images.
STAC~\cite{sohn2020simple} uses strong data augmentation for self-training.

Recently, pseudo-labeling-based methods have shown promising results on several benchmark datasets, which are attributed to a stronger teacher model trained by, e.g., a weighted EMA ensemble~\cite{liu2021unbiased,tang2021humble,yang2021interactive,xu2021end,zhang2022semi,chen2022dense,chen2022label}, a data ensemble~\cite{tang2021humble}, or advanced data augmentation~\cite{zhou2021instant,tang2021humble}.
To overcome the confirmation bias, Unbiased Teacher~\cite{liu2021unbiased} employs focal loss~\cite{lin2017focal} to reduce the weights on overconfident pseudo labels, while others use uncertainty modeling~\cite{wang2021data} or co-training~\cite{zhou2021instant} as the countermeasure.
Li, \textit{et al.}~\cite{li2022rethinking} propose dynamic thresholding for each class based on both localization and classification confidence.
LabelMatch~\cite{chen2022label} introduces a re-distribution mean teacher based on the 
 KL divergence distribution between teacher and student models.
DSL~\cite{chen2022dense} assigns pixel-wise pseudo-labels for anchor-free detectors.
Unbiased Teacherv2~\cite{liu2022unbiased} introduces a new pseudo-labeling mechanism based on the relative uncertainties of teacher and student models.

It is worth noting that most existing methods are designed for class-balanced datasets like MS COCO~\cite{lin2014microsoft}, while their capabilities to handle long-tailed datasets like LVIS~\cite{gupta2019lvis} have been largely under-studied---to our knowledge, \emph{none of existing research has specifically investigated long-tailed object detection in the SSL setting}. Instead, the majority of existing SSL algorithms are evaluated on class-balanced datasets~\cite{jeong2019consistency,sohn2020simple,liu2021unbiased,xu2021end}.
Our work takes the first step toward a unified approach to solving unlabeled data and the long-tailed object detection problem, which we hope to inspire more work to tackle this challenging setting.

\smallsec{Long-tailed Object Detection}
Though object detection has witnessed significant progress in recent years~\cite{ren2015faster,lin2017focal,cai2019cascade,tian2019fcos,carion2020end,sun2021sparse}, how to deal with the long-tailed problem remains an open question~\cite{zhang2021deep}.
Most existing methods fall into two groups: data re-sampling~\cite{gupta2019lvis,shen2016relay,hu2020learning,wu2020forest} and loss re-weighting~\cite{tan2020equalization,ren2020balanced,wang2021adaptive,tan2020equalization2,zhang2021distribution,wang2020seesaw,feng2021exploring,chang2021image,zhou2021probabilistic,li2022equalized,he2022relieving}.
Some recent works~\cite{zang2021fasa,li2021metasaug,ghiasi2021simple} suggest that data augmentation is useful for long-tailed recognition.
In terms of data re-sampling, Repeated Factor Sampling (RFS)~\cite{gupta2019lvis} assigns high sampling rates to images of rare classes.
A couple of studies~\cite{li2020overcoming,wang2020devil} have suggested using different sampling schemes in decoupled training stages.
When it comes to data re-weighting, a representative method is equalization loss~\cite{tan2020equalization,tan2020equalization2}, which raises the weights for rare classes based on inverse class frequency.
Seesaw Loss~\cite{wang2020seesaw} automatically adjusts class-specific loss weights based on a statistical ratio between the positive and negative gradients computed for each class.
MosaicOS~\cite{zhang2021mosaicos} is one of the early studies that uses weakly-supervised learning to help long-tailed detection. Their study assumes the availability of weakly-annotated class labels. In contrast, we take a pure semi-supervised setting without assuming any annotations in the unlabeled set.
In our work, we first investigate how to exploit unlabeled data to improve the performance of detectors trained on long-tailed datasets.

\smallsec{Semi-Supervised Learning (SSL)}
Numerous SSL methods are based on consistency learning~\cite{sajjadi2016regularization,berthelot2019mixmatch,berthelot2020remixmatch,sohn2020fixmatch,zheng2022simmatch,yang2022class}, which forces a model's predictions on two different views of the same instance to be similar.
Recent state-of-the-art consistency learning methods like MixMatch~\cite{berthelot2019mixmatch}, UDA~\cite{xie2020unsupervised} and FixMatch~\cite{sohn2020fixmatch} introduce strong data augmentations~\cite{xie2020unsupervised} to the learning paradigm---they use predictions on weakly augmented images as the target to train the model to produce similar outputs given the strongly augmented views of the same images.

Another research direction related to our work is pseudo-labeling~\cite{bachman2014learning,lee2013pseudo,iscen2019label,xie2020self,oh2022distribution}, which is typically based on a teacher-student architecture: a teacher model's predictions are used as the target to train a student model.
The teacher model can be either a pretrained model~\cite{sohn2020fixmatch} or an exponential moving average of the student model~\cite{rasmus2015semi,laine2017temporal,tarvainen2017mean,liu2021unbiased}.
Some studies~\cite{arazo2020pseudo} have also demonstrated that using the student model being trained to produce the target can reach decent performance---the trick is to inject strong noise to the student model, such as applying strong data augmentations to the input~\cite{sohn2020fixmatch}.

A common issue encountered in pseudo-labeling methods is confirmation bias~\cite{arazo2020pseudo}, which is caused by a constant feed of incorrect pseudo labels with high confidence to the model. And such a vicious cycle would reinforce since the model will become increasingly inaccurate and subsequently provide more erroneous pseudo labels.
To mitigate the issue of confirmation bias, existing methods have tried using an uncertainty-based metric~\cite{rizve2021defense} to modulate the confidence threshold or using the co-training framework~\cite{han2018co,qiao2018deep} that simultaneously trains two neural networks each giving pseudo labels to the other.
In this work, to prevent each detection head from overfitting its own prediction errors, the pseudo labels to train each detection head are formed by the ensemble predictions of multiple detection heads. This strategy is new in the literature.

It is worth noting that most aforementioned algorithms are evaluated on class-balanced datasets while only very few recent works apply SSL for long-tailed image classification~\cite{hyun2020class, kim2020distribution, yang2020rethinking, wei2021crest,lee2021abc,fan2022cossl,oh2022distribution} or semantic segmentation~\cite{he2021re,hu2021semi}.
The detection task requires predicting both the class labels and object locations, which is much harder than the classification-only task.
The pseudo-labeling-based semi-supervised methods are unable to predict high-quality pseudo labels for detection task as accurately as for classification task, in the presence of class imbalance. This motivates us to improve the pseudo-labeling quality for semi-supervised and long-tailed detection using a cascade mechanism.

\section{Our Approach: CascadeMatch}

\begin{figure*}[!htb]
\centering
\includegraphics[width=1.00\textwidth]{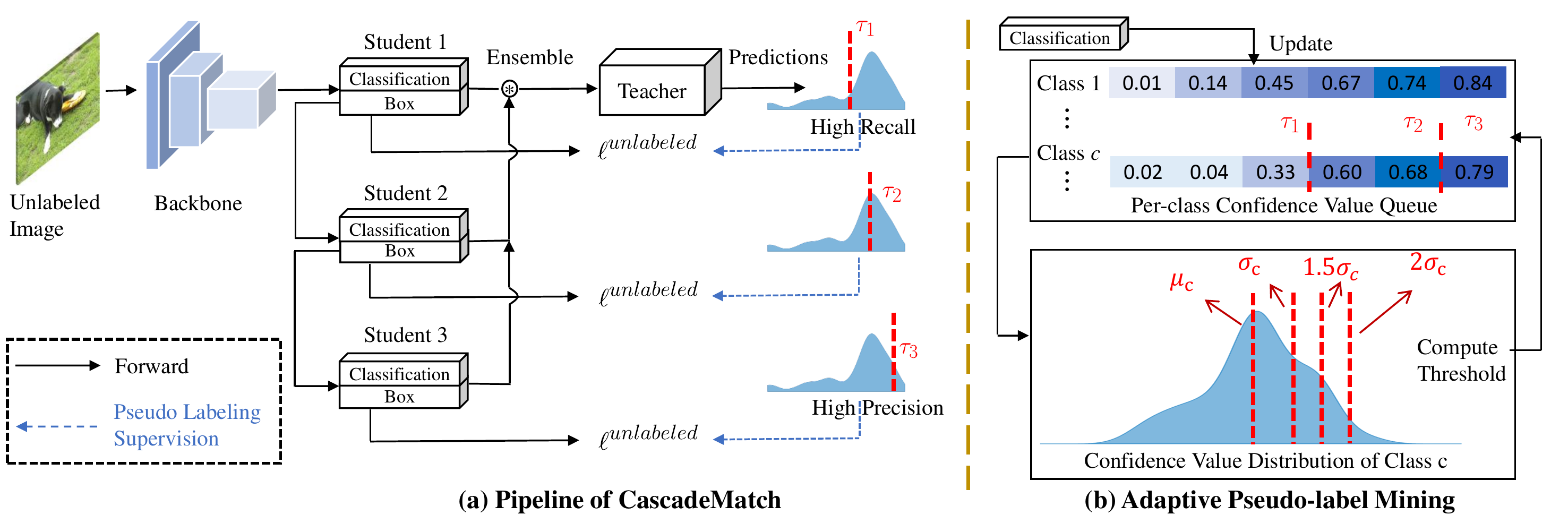}
\caption{The pipeline of our approach. \textbf{(a)}: Overview of CascadeMatch's cascade pseudo-labeling module. The supervision signal for unlabeled data corresponds to the ensembled pseudo label. Confidence thresholds, $\{\tau_k\}_{k\in{1,\dots,3}}$, are independently computed for each stage via our adaptive pseudo-label mining module.
\textbf{(b)}: Computation of the adaptive pseudo-label mining module. The classification confidence values predicted for each class $c \in \{1, \dots, C\}$ on labeled proposals are aggregated in the per-class queue. For class $c$, the confidence value distribution is estimated where the mean $\mu_c$ and the standard deviation $\sigma_c$ are used to determine the class-specific threshold$\tau_k^c$ at the $k$-th cascade stage.
}
\label{fig:cascade}
\end{figure*}

\smallsec{Problem Definition}
Given a labeled dataset $\mathcal{D}_l = \{ (\bm{x}, y^*, \bm{b}^*) \}$ with $\bm{x}$, $y^*$ and $\bm{b}^*$ denoting image, label and bounding box, respectively,\footnote{For simplicity, we use a single proposal in our formulations, which can be easily extended to a batch of proposals.} and an unlabeled dataset $\mathcal{D}_u = \{\bm{x}\}$, the goal is to learn a robust object detector using both $\mathcal{D}_l$ and $\mathcal{D}_u$. We further consider the issue of long-tailed distribution~\cite{gupta2019lvis}, which is common in real-world data but have been largely unexplored in existing semi-supervised object detection methods. More specifically, let $n_i$ and $n_j$ denote the number of images for class $i$ and $j$ respectively, and assume $i$ is a frequent class while $j$ is a rare class. In a long-tailed scenario, we might have $n_i \gg n_j$.

\smallsec{An Overview}
A brief overview of the main paradigm of our proposed CascadeMatch is illustrated in Figure~\ref{fig:cascade}. CascadeMatch features a cascade pseudo-labeling~(CPL) design and an adaptive pseudo-label mining~(APM) mechanism. The former aims to generate pseudo-labels and filter out low-quality labels in a cascade fashion to improve the trade-off between precision and recall, while the latter aims to automate threshold tuning. 
CascadeMatch only modifies a detector's head structure and thus can be seen as a plug-and-play module that fits into most existing object detectors including the popular anchor-based R-CNN series like Cascade R-CNN~\cite{cai2019cascade} or more recent end-to-end detectors like Sparse R-CNN~\cite{sun2021sparse}. CascadeMatch can also take either CNNs~\cite{he2016deep} or Transformers~\cite{liu2021swin} as the backbone.

\smallsec{Discussion}
A cascade structure benefits from the ``divide and conquer'' concept, where each stage is dedicated to a specific sub-task. This notion of cascading has been found practical and useful in many computer vision systems. 
For the detection task, finding an accurate IoU threshold to separate the \emph{positive} and \emph{negative} region proposals is impossible. To allow a better precision-recall trade-off, Cascade R-CNN uses the cascade structure to progressively increase the IoU threshold for different stages.
Recall that pseudo labeling faces a similar dilemma in pinpointing a single confidence threshold to separate the valid \emph{pseudo-labels} and noisy \emph{background} region proposals. It is thus natural for CascadeMatch to use the cascade structure with a set of progressive confidence thresholds. 
Note that the confidence threshold of CascadeMatch is class-specific and self-adaptive. We will provide the details in Section~\ref{sec:method;subsec:threshold}.

Below we provide the technical details of the two key components in CascadeMatch, namely cascade pseudo-labeling (Section~\ref{sec:method;subsec:cascade}) and adaptive pseudo-label mining (Section~\ref{sec:method;subsec:threshold}). For clarity, in Section~\ref{sec:method;subsec:cascade} we first present CascadeMatch in an anchor-based framework and later explain the modifications needed for an end-to-end detector.

\subsection{Cascade Pseudo-Labeling}
\label{sec:method;subsec:cascade}

\smallsec{Model Architecture}
For an anchor-based framework~\cite{ren2015faster,cai2019cascade}, the CascadeMatch-based detector starts with a CNN as the backbone for feature extraction, e.g., ResNet50~\cite{he2016deep}, which is then followed by a region proposal network (RPN)~\cite{ren2015faster} for generating object proposals. See Figure~\ref{fig:cascade}(a) for the architecture.

The detector has $K$ heads following the Cascade R-CNN~\cite{cai2019cascade} pipeline.
The parameter $K$ controls the trade-off between performance and efficiency, which can be adjusted by practitioners based on their needs. Increasing the number of heads will improve the performance at the cost of speed. In the paper, we followed previous cascade methods~\cite{cai2019cascade,sun2021sparse} to use $K=3$ heads. We will provide the ablation studies of varying the value of $K$ in Table~\ref{tab:K} of Section~\ref{sec:abla_studies}.
Formally, given an image $\bm{x}$, the first-stage detection head predicts for an object proposal $\bm{b}_0$ (generated by the RPN) a class probability distribution $p_1(y | \bm{x}, \bm{b}_0)$ and the bounding box offsets $\bm{b}_1$. Then, the second-stage detection head predicts another probability $p_2(y | \bm{x}, \bm{b}_1)$ using the refined bounding box from the first stage;\footnote{With a slight abuse of notation, $\bm{b}_1$ in $p_2(y | \bm{x}, \bm{b}_1)$ contains the complete coordinates of the bounding box rather than the regressed offsets.} and so on and so forth.

\smallsec{Labeled Losses}
With labeled data $\mathcal{D}_l = \{ (\bm{x}, y^*, \bm{b}^*) \}$, we train each detection head using the classification loss $\operatorname{Cls(\cdot, \cdot)}$ (for proposal classification) and the bounding box regression loss $\operatorname{Reg(\cdot, \cdot)}$~\cite{ren2015faster}. Formally, we have
\begin{align}
\ell_{cls}^{labeled} &= \sum_{(\bm{x}, y^*) \sim \mathcal{D}_l} \sum_{k=1}^K \operatorname{Cls}(y^*, p_k(y|\bm{x}, \bm{b}_{k-1})), \label{eq:loss_labeled_cls} \\
\ell_{reg}^{labeled} &= \sum_{(\bm{x}, \bm{b}^*) \sim \mathcal{D}_l} \sum_{k=1}^K \operatorname{Reg}(\bm{b}^*, \bm{b}_k). \label{eq:loss_labeled_reg}
\end{align}

\smallsec{Unlabeled Losses}
To cope with unlabeled images, we adopt a pseudo-labeling approach with a teacher-student architecture where the teacher's estimations on unlabeled data are given to the student as supervision. Such a paradigm has been widely used in previous semi-supervised methods~\cite{sohn2020simple,liu2021unbiased,zhou2021instant,tang2021proposal,tang2021humble,wang2021data}. Different from previous methods, we focus on tackling the confirmation bias issue~\cite{arazo2020pseudo} when designing our architecture. We observe that the ensemble predictions are more accurate than using each individual prediction~(please refer to Table~\ref{tab:cofirmation_bias} of Section~\ref{sec:abla_studies} for more details), so we use the ensemble predictions from all detection heads as the teacher supervision signal (teacher module in Figure.~\ref{fig:cascade} (\textbf{a})). Formally, given an unlabeled image $\bm{x} \sim \mathcal{D}_u$, the ensemble prediction $p_{t}$ is computed as
\begin{equation}
p_{t} = \frac{1}{K} \sum_{k=1}^K p_k(y | \bm{x}, \bm{b}_{k-1}) \quad \text{and} \quad \bm{b}_{t} = \frac{1}{K} \sum_{k=1}^K \bm{b}_k,
\end{equation}
where $K$ is the number of heads. Let $q_{t} = \max(p_{t})$ be the confidence and $\hat{q}_{t} = \arg\max(p_{t})$ the pseudo label, we compute the classification loss and the bounding box regression loss for unlabeled data using
\begin{align}
\ell_{cls}^{unlabeled} &= \sum_{\bm{x} \sim \mathcal{D}_u} \sum_{k=1}^K \mathds{1}(q_{t} \geq \tau_k^{\hat{q}_{t}}) \operatorname{Cls}(\hat{q}_{t}, p_k(y|\bm{x}, \bm{b}_{k-1})), \label{eq:loss_unlabeled_cls} \\
\ell_{reg}^{unlabeled} &= \sum_{\bm{x} \sim \mathcal{D}_u} \sum_{k=1}^K \mathds{1}(q_{t} \geq \tau_k^{\hat{q}_{t}}) \operatorname{Reg}(\bm{b}_{t}, \bm{b}_k), \label{eq:loss_unlabeled_reg}
\end{align}
where $\tau_k^{\hat{q}_{t}}$ is a self-adaptive confidence threshold specific to class $\hat{q}_{t}$. We detail the design of class-specific self-adaptive thresholds in Section~\ref{sec:method;subsec:threshold}.

\smallsec{Training}
Similar to most region-based object detectors, our CascadeMatch model is learned using four losses: a region-of-interest (ROI) classification loss $\ell_{cls}^{roi} = \ell_{cls}^{labeled} + \lambda^{u} \cdot \ell_{cls}^{unlabeled}$, an ROI regression loss $\ell_{reg}^{roi} = \ell_{reg}^{labeled} + \lambda^{u} \cdot \ell_{reg}^{unlabeled}$, and two other losses for the RPN, i.e.,~the objectness classification loss $\ell_{cls}^{rpn}$ and the proposal regression loss $\ell_{reg}^{rpn}$, as defined in~\cite{ren2015faster}.
The loss parameter $\lambda^{u}$ controls the weight between the supervised term $\ell_{cls}^{l}$ and the unsupervised term $\ell_{cls}^{u}$.
By default, we set the unsupervised loss weight $\lambda_{u}=1.0$.

\smallsec{Transfer to End-to-End Object Detector}
CascadeMatch is readily applicable to an end-to-end detector. We use Sparse R-CNN~\cite{sun2021sparse} as an example. Two main modifications are required: 1) Since region proposals are learned from a set of embedding queries as in DETR~\cite{carion2020end}, we do not need an RPN and the RPN loss $\ell^{rpn}$; 2) The classification loss is replaced by the focal loss~\cite{lin2017focal} while the regression loss is replaced by L1 and GIoU loss~\cite{rezatofighi2019generalized}. \emph{We show the universality of CascadeMatch on anchor-based detector (i.e., Cascade R-CNN) and an end-to-end detector (\ie, Sparse R-CNN) in the experiments}, see Table~\ref{tab:mainresults_lvis}.

\subsection{Adaptive Pseudo-label Mining} \label{sec:method;subsec:threshold}
\begin{figure*}[t]
\centering
\begin{minipage}{0.495\textwidth}
    \centering
    \includegraphics[width=\textwidth]{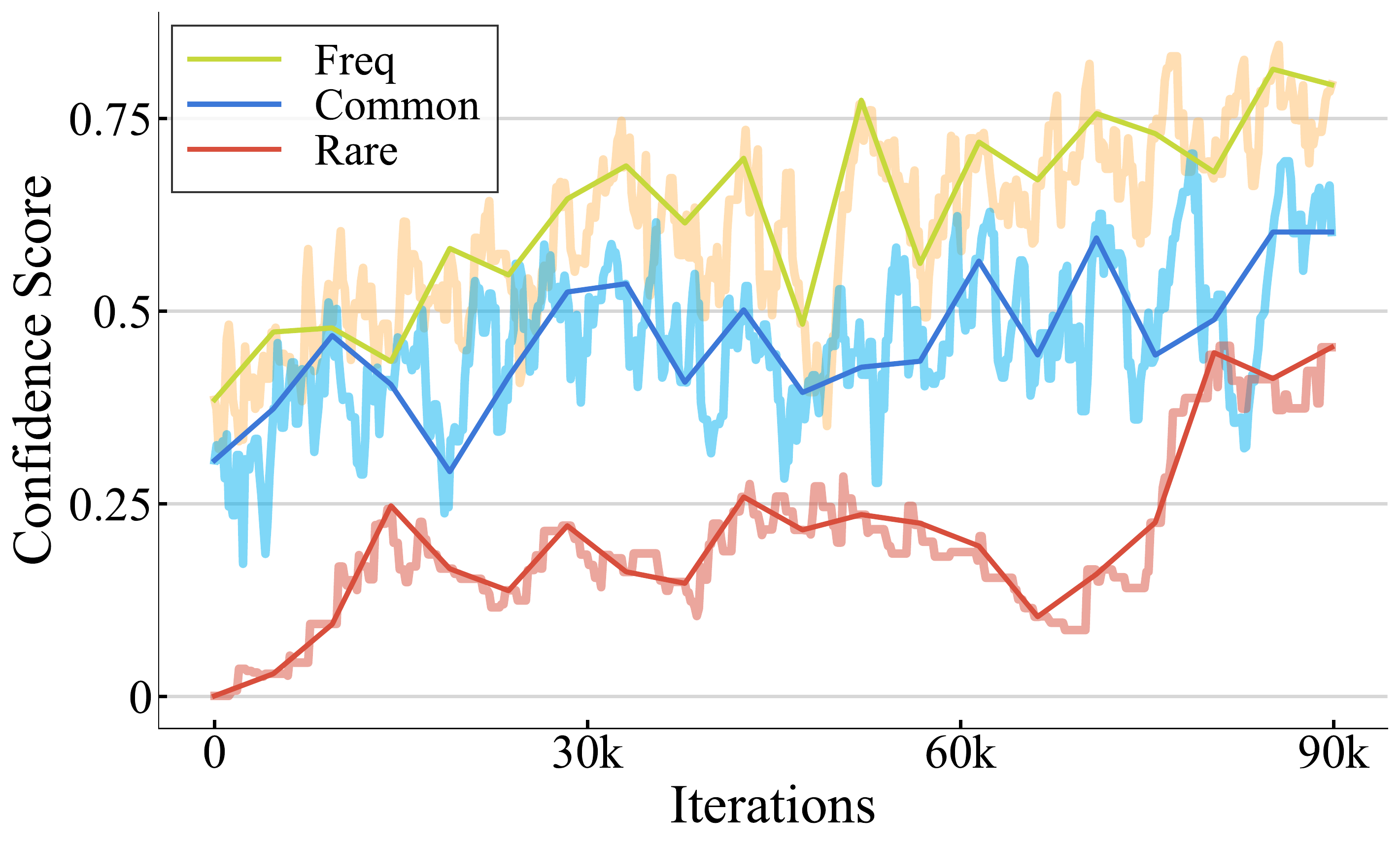}
\end{minipage}
\begin{minipage}{0.495\textwidth}
    \centering
    \includegraphics[width=\textwidth]{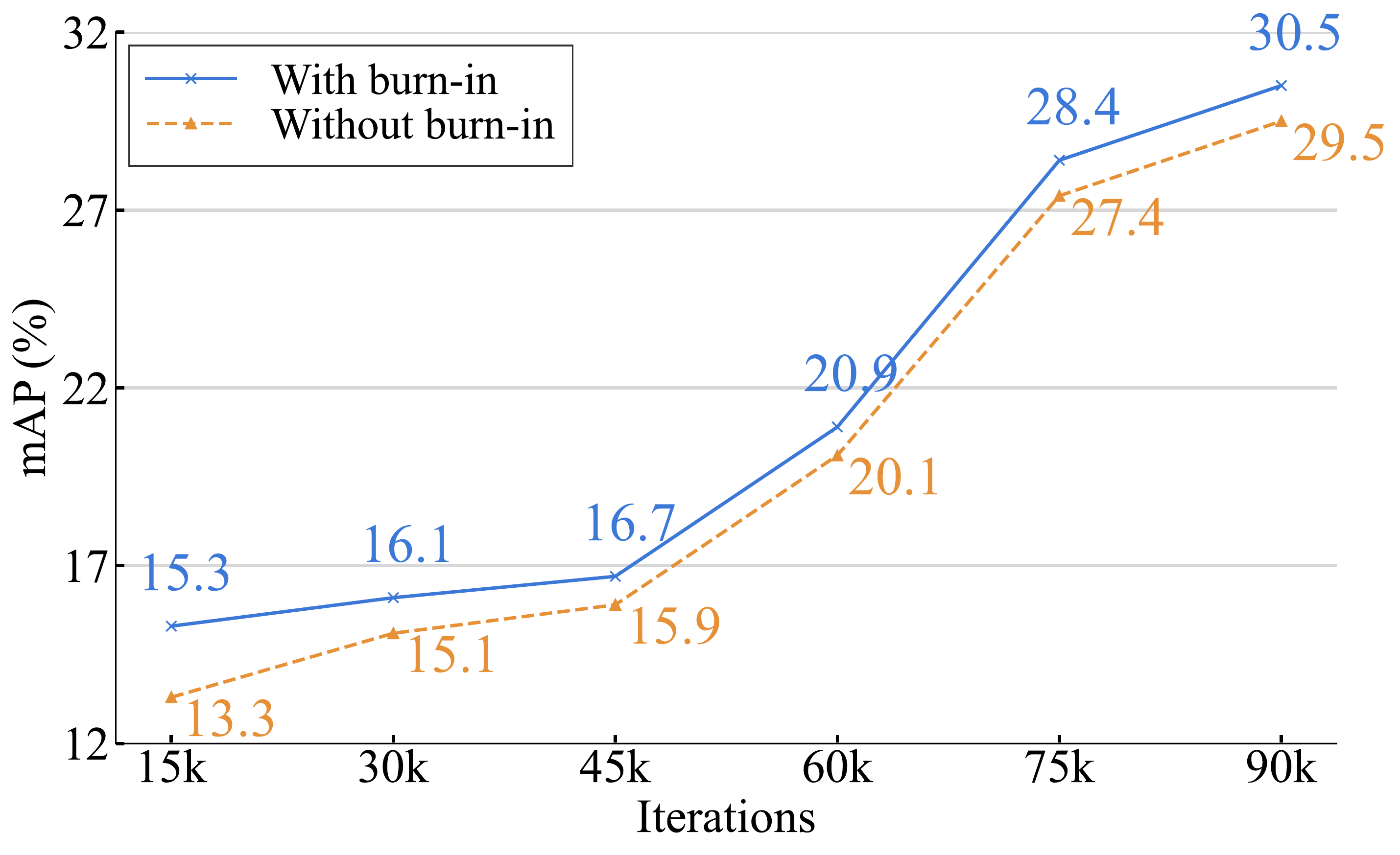}
\end{minipage}
\caption{\textbf{(a)} Visualization of predictive confidence scores throughout training. We find that the predicted scores have the increasing tendency, which motivates us to propose the Adaptive Pseudo-label Mining~(APM) module that using dynamic thresholds. \textbf{(b)} Impact of the burn-in stage. Clearly, the burn-in stage improves the performance.}
\label{fig:burn-in}
\end{figure*}

Determining a confidence threshold for pseudo labels is a non-trivial task, not to mention that each class requires a specific threshold to overcome the class-imbalance issue---many-shot classes may need a higher threshold while few-shot classes may favor a lower threshold. Moreover, predictive confidence typically increases as the model observes more data (see Figure~\ref{fig:burn-in})~(a), and therefore, dynamic thresholds are more desirable.

To solve the aforementioned problems, we propose an Adaptive Pseudo-label Mining (APM) module, which is an \emph{automatic} selection mechanism for predicted pseudo-labels.
Specifically, at each iteration, we first aggregate the ensemble predictions made on each ground-truth class using the labeled proposals (see Figure~\ref{fig:cascade}(a)), and then select a threshold such that a certain percentage of the confidence values can pass through.
The challenge lies in how to select the threshold with minimal human intervention.
We automate the selection process by (1) computing the mean $\mu_c$ and the standard deviation $\sigma_c$ based on the confidence values for each class, and (2) setting the class-specific threshold $\tau_k^c$ for stage-$k$ as $\tau_k^c = \mu_c + \sigma_c * \epsilon_k$. An illustration is shown in Figure~\ref{fig:cascade}(b). 

The formulation above is simple but meaningful. In particular, since the predictive confidence values for each class are updated every iteration, the mean $\mu_c$ will increase gradually, which naturally makes $\tau_k^c$ self-adaptive to the learning process without extra designs. By increasing $\epsilon_k$ moderately in different stages, we maintain the progressive pattern of confidence threshold for different stages~(\eg,  $\tau_1 < \tau_2 < \cdots < \tau_K$) for any class. In this work, we choose $\epsilon_k \in \{1, 1.5, 2\}$ for the three stages. The ablation study is provided in Table~\ref{tab:epsilon} of Section~\ref{sec:abla_studies}. In the experiments, we show that the progressive design is useful to control the precision and recall trade-off.

\section{Experiments}
\label{sec:experiments}

\smallsec{Datasets}
We evaluate our approach on two \emph{long-tailed} object detection datasets: LVIS v1.0~\cite{gupta2019lvis} and COCO-LT~\cite{wang2020devil}.
LVIS v1.0 widely serves as a testbed for the long-tailed object detection task~\cite{tan2020equalization,tan2020equalization2,li2020overcoming,wang2020devil,hu2020learning,zhang2021distribution,wang2020seesaw,feng2021exploring,chang2021image,zhou2021probabilistic}.
Three class groups are defined in LVIS v1.0: rare [1, 10), common [10, 100), and frequent [100, -) based on the number of images that contain at least one instance of the corresponding class.
COCO-LT~\cite{wang2020devil} is used to demonstrate the generalizability of our approach. Similarly, COCO-LT defines four class groups with the following ranges: [1, 20), [20, 400), [400, 8000), and [8000, -).
For both LVIS and COCO-LT, we use the MS-COCO 2017 \textit{unlabeled} set as the unlabeled dataset, which contains 123,403 images in total and has a labeled-to-unlabeled ratio of roughly $1:1$.

\smallsec{Metrics}
We adopt the recently proposed Fixed AP (denoted by \fixedap) metric~\cite{dave2021evaluating}, which does not restrict the number of predictions per image and can better characterize the long-tailed object detection performance. Following Dave et al.~\cite{dave2021evaluating}, we adopt the following notations for the metrics of different class groups: \fixedapr for rare classes, \fixedapc for common classes, and \fixedapf for frequent classes. For COCO-LT dataset, the symbols \apone, \aptwo, \apthree and \apfour correspond to the bins of $[1, 20)$, $[20, 400)$, $[400, 8000)$ and $[8000, -)$ (\ie, number of training instances).

\smallsec{Implementation Details}
\begin{table}[t]
\begin{minipage}{.46\textwidth}
  \centering
  \caption{
  List of hyper-parameters used for different detectors.
  }
  \label{tab:hyperparameter}
  \footnotesize
  \scalebox{1.0}{
  \tablestyle{1pt}{1.0}
    \begin{tabular}
    {lcx{30}}
    \toprule
    Hyper-parameter  & Detector & Value \\
    \cmidrule(r){1-1}
    \cmidrule(r){2-2}
    \cmidrule(r){3-3}
    Optimizer & \multirow{3}*{Cascade R-CNN} & SGD \\
    Learning Rate & ~ & 0.01 \\
    Weight Decay & ~ & 0.0001 \\
    \midrule
    Optimizer & \multirow{3}*{Sparse R-CNN} & AdamW \\
    Learning Rate & ~ & 0.000025 \\
    Weight Decay & ~ & 0.0001 \\
    \midrule
    Input Image Size & \multirow{3}*{Both} & $[1333, 800]$ \\
    Batch Size for Labeled Data & & 16  \\
    Batch Size for Unlabeled Data & & 16 \\
    \bottomrule
    \end{tabular}}
\end{minipage}
\end{table}
For the anchor-based detector, we employ the two-stage detector, Cascade R-CNN~\cite{cai2019cascade} with the FPN~\cite{lin2017feature} neck. ResNet50~\cite{he2016deep} pre-trained from ImageNet is used as the CNN backbone. For the end-to-end detector, we adopt Sparse R-CNN~\cite{sun2021sparse} with the Pyramid Vision Transformer (PvT)~\cite{wang2021pyramid} encoder. All settings for the parameters, such as learning rate, are kept the same as previous work~\cite{liu2021unbiased}. We list the value of our used hyper-parameters in Table~\ref{tab:hyperparameter}. All models are trained with the standard SGD optimizer on 8 GPUs. Similar to previous methods~\cite{sohn2020simple,liu2021unbiased,tang2021humble}, we also have a ``burn-in'' stage to stabilize training. Specifically, we pre-train the detector using labeled data first for several iterations, and then include unlabeled data in the training process.

\subsection{Ablation Studies} \label{sec:abla_studies}

\begin{table*}[t!]
\begin{minipage}{.5\textwidth}
\centering
\caption{\small{
	Ablation studies on 1) cascade pseudo-labeling (CPL) and 2) adaptive pseudo-label mining (APM).
	The top row refers to the supervised learning baseline without using the unlabeled data.
}}
\label{tab:ablation_lvis}
\scalebox{1.0}{\tablestyle{4pt}{1.0}
\begin{tabular}{x{15}x{15}x{25}x{25}x{25}x{25}}
    \midrule
    CPL & APM & \fixedap & \fixedapr & \fixedapc & \fixedapf \\
    \cmidrule(r){1-2}
    \cmidrule(r){3-6}
    \gxmark& \gxmark & 26.3 & 19.7 & 25.3 & 30.3 \\
    \cmark & \gxmark & 30.1 & 21.9 & 29.3 & 34.5 \\ 
    \gxmark& \cmark  & 28.9 & 22.5 & 27.9 & 32.8 \\
    \cmark & \cmark  & \bf{30.5} & \bf{23.1} & \bf{29.7} & \bf{34.7} \\
    \bottomrule
\end{tabular}}
\end{minipage}
\hspace*{+2mm}
\begin{minipage}{.38\textwidth}
\centering
\caption{
	\small{Ablation study on the selection of the confidence parameter $\epsilon$. We observe that the $\epsilon$ works the best with progressive values ($\epsilon_{1} < \epsilon_{2} < \epsilon_{3}$).}
}
\label{tab:epsilon}
	\centering
	\scalebox{1.0}{
	\tablestyle{6pt}{1.0}
	\begin{tabular}{x{6}x{6}x{6}x{20}x{20}x{20}x{20}}
    \toprule
    $\epsilon_{1}$ & $\epsilon_{2}$ & $\epsilon_{3}$ & \fixedap & \fixedapr & \fixedapc & \fixedapf \\
    \cmidrule(r){1-3}
    \cmidrule(r){4-7}
    0.0 & 0.0 & 0.0 & 29.8 & 21.7 & 29.1 & 34.1 \\
    0.0 & 1.0 & 2.0 & 30.2 & \textbf{23.3} & 29.2 & 34.3 \\
    1.0 & 2.0 & 3.0 & 30.3 & 22.6 & 29.5 & 34.4 \\
    \cmidrule(r){1-3}
    \cmidrule(r){4-7}
    1.0 & 1.5 & 2.0 & \GrayBG{\bf{30.5}} & \GrayBG{23.1} & \GrayBG{\textbf{29.7}} & \GrayBG{\textbf{34.7}} \\
    \bottomrule
    \end{tabular}}
\end{minipage}
\end{table*}

\begin{table*}[t]
\centering
\begin{minipage}[b]{.30\linewidth}
\caption{
	\small{Ablation study on the number of detector heads $K$. We also report the training time~(seconds) per iteration in the last column. }
}
\label{tab:K}
	\centering
	\scalebox{0.95}{
	\tablestyle{4pt}{1.0}
	\begin{tabular}{x{6}x{20}x{20}x{20}x{20}x{20}}
    \toprule
    $K$ & \fixedap & \fixedapr & \fixedapc & \fixedapf & $\operatorname{T}_{\mathrm{train}}$ \\
    \cmidrule(r){1-1}
    \cmidrule(r){2-5}
    \cmidrule(r){6-6}
   1 & 26.4 & 20.4 & 26.6 & 28.9 & 0.36\\
   2 & 28.0 & 21.4 & 27.1 & 31.9 & 0.42 \\
   3 & \GrayBG{\bf{30.5}} & \GrayBG{\textbf{23.1}} & \GrayBG{\textbf{29.7}} & \GrayBG{34.7} & 0.47 \\
   4 & 30.0 & 22.1 & 29.2 & 34.6 & 0.59 \\
   5 & 29.9 & 21.2 & 29.0 & \textbf{34.9} & 0.72 \\
   \bottomrule
   \end{tabular}}
\end{minipage}
\hspace*{+2mm}
\begin{minipage}[b]{.30\linewidth}
\caption{
	\small{Comparison of pseudo-label accuracy. The ensemble results is more accurate than each single head. See Figure~\ref{fig:vis_pseudo} for visualization.}
}
\label{tab:cofirmation_bias}
	\centering
    \scalebox{1.0}{\tablestyle{6pt}{1.0}\begin{tabular}{cccc}
    \toprule
    Iter. & 60k & 120k & 180k \\
    \cmidrule(r){1-1}
    \cmidrule(r){2-4}
    Head 0 & 32.8 & 51.5 & 67.3 \\
    Head 1 & 50.5 & 62.4 & 73.2 \\
    Head 2 & 55.1 & 71.0 & 84.1 \\
    Ensemble & \GrayBG{\textbf{66.4}} & \GrayBG{\textbf{79.5}} & \GrayBG{\textbf{88.9}} \\
    \bottomrule
    \end{tabular}}
\end{minipage}
\hspace*{+1mm}
\begin{minipage}[b]{.30\linewidth}
\caption{
 \small{Ablation study on the loss function weight balancing parameter $\ell_{cls}^{u}$. We select $\ell_{cls}^{u}=1.0$ that works the best.}
}
\label{tab:unsup_weight}
    \centering
    \scalebox{1.00}{\tablestyle{3pt}{1.0}
    \begin{tabular}{lx{30}x{30}x{30}x{30}}
     \toprule
    $\lambda^{u}$  & \fixedap & \fixedapr & \fixedapc & \fixedapf \\
    \cmidrule(r){1-1}
    \cmidrule(r){2-5}
    0.5 & 30.0 & 20.9 & 28.2 & 36.1 \\
    1.5 & 29.9 & 21.2 & 28.3 & 35.6 \\
    1.0 & \GrayBG{\bf{30.5}} & \GrayBG{\bf{21.4}} & \GrayBG{\bf{28.9}} & \GrayBG{\bf{36.4}} \\
    2.0 & 29.4 & 20.4 & 27.9 & 35.1 \\
    \bottomrule
    \end{tabular}}
\end{minipage}
\end{table*}

Before discussing the main results of long-tailed and semi-supervised object detection, we investigate the effects of the two key components of CascadeMatch, \ie, the cascade pseudo-labeling (CPL) and adaptive pseudo-label mining (APM), as well as some hyper-parameters. The experiments are conducted on the LVIS v1.0 \emph{validation} dataset.

\smallsec{Cascade Pseudo-Labeling}
The results are detailed in Table~\ref{tab:ablation_lvis}. We first examine the effect of the cascade pseudo-labeling module. The top row contains the results of the supervised baseline, while the second row corresponds to the combination of the baseline and CPL. We observe that CPL clearly improves upon the baseline. Notably, CPL improves the performance in all groups: +2.2 for the rare classes, +4.0 for the common classes, and +4.2 for the frequent classes.

\smallsec{Adaptive Pseudo-label Mining}
We then examine the effectiveness of APM. By comparing the first and third rows in Table~\ref{tab:ablation_lvis}, we can conclude that APM alone is also beneficial to the performance, yielding clear gains of $2.8$ \fixedapr and $2.6$ \fixedapc. Finally, by combining CPL and APM (the last row), the performance can be further boosted, suggesting that the two modules are complementary to each other for long-tailed and semi-supervised object detection. We observe that CPL+APM brings a non-trivial improvement of 1.2$\%$ to the rare classes compared with using CPL only. The predictions on rare classes often have smaller confidence so the class-specific design in APM is essential for handling the long-tailed issue.

\smallsec{Hyper-parameter $\epsilon_k$}
As discussed in Section~\ref{sec:method;subsec:threshold}, our confidence thresholds $\tau_k$ are adaptively adjusted and governed by a hyper-parameter $\epsilon_{k}$. In Table~\ref{tab:epsilon}, we show the effects of using different values for $\epsilon_k$ to update the per-class thresholds. Overall, the performance is insensitive to different values of $\epsilon_k$, with $\epsilon_k = \{1.0, 1.5, 2.0\}$ achieving the best performance.

\smallsec{Hyper-parameter $K$}
The parameter $K$ denotes the number of detection heads.
We try different values of $K$, and the results are shown in Table~\ref{tab:K}.
We observe that from $k=1$ to $3$, increasing the number of heads will improve the overall performance at the cost of training speed.
The performance of rare and common classes will drop if we continue to increase the $k$ from $3$ to $4$ or $5$, probably due to the over-fitting and undesired memorizing effects of few-shot classes as we increase the model capacity.
In this study, we choose to follow previous cascade methods~\cite{cai2019cascade} that use $K=3$ heads.

\smallsec{Confirmation Bias}
Recall that we use the ensemble teacher to train each detection head instead of using each individual prediction to mitigate confirmation bias. To understand how our design tackles the problem, we print the pseudo-label accuracy obtained during training for each detection head and their ensemble. Specifically, we use 30\% of the LVIS training set as the labeled set and the remaining 70\% as the unlabeled set. Note that the annotations for the unlabeled data are used only to calculate the pseudo-label accuracy. The results obtained at the 60k-th, 120k-th and 180k-th iteration are shown in Table~\ref{tab:cofirmation_bias}. It is clear that the pseudo-label accuracy numbers for individual heads are consistently lower than that of the ensemble throughout the course of training, confirming that using ensemble predictions is the optimal choice.

\smallsec{Hyper-parameter $\lambda^{u}$}
To examine the effect of unsupervised loss weights $\lambda^{u}$, we vary the unsupervised loss weight $\lambda_{u}$ from 0.5 to 2.0 on LVIS~\cite{gupta2019lvis} dataset.
As shown in Table~\ref{tab:unsup_weight}, we observe that the model performs best with our default choice $\lambda_{u}=1.0$.

\smallsec{Burn-in Stage}
As mentioned at the beginning of Section~\ref{sec:experiments}, we set a `burn-in' stage to pre-train the detector on the labeled data before training on unlabeled data.
Similar to previous works~\cite{sohn2020simple,liu2021unbiased,tang2021humble}, such a `burn-in' stage is used to stabilize initialization results in the early stage of training.
In Figure~\ref{fig:burn-in}~(b), we provide the mAP comparison of the CascadeMatch with and without the burn-in stage during the training.
We observed that the model achieves higher mAP in the early stage with the burn-in stage and converges into better endpoints compared with the counterparts.

\subsection{Main Results}
\begin{table*}[t]
  \caption{Comparisons of mAP against the supervised baseline and different semi-supervised methods on LVIS v1.0 \textit{validation} set
  We select two different frameworks: Cascade R-CNN~\cite{cai2019cascade} and Sparse R-CNN~\cite{sun2021sparse} with different backbones as the supervised baseline.
  The symbols \fixedapr, \fixedapc, and \fixedapf refer to the Fixed mAP~\cite{dave2021evaluating} of overall, rare, common, and frequent class groups.
  The `12e' and `30e' schedules refer to 12 and 30 epochs, respectively.
  We report the average results over three runs with different random seeds.
  }
  \label{tab:mainresults_lvis}
  \centering
  \footnotesize
    \scalebox{1.0}{\tablestyle{2pt}{1.0}
    \begin{tabular}{lx{70}x{50}x{30}x{40}x{40}x{40}x{40}}
    \toprule
    Method  & Framework & Backbone & Schedule & \fixedap & \fixedapr & \fixedapc & \fixedapf \\
    \cmidrule(r){1-1}
    \cmidrule(r){2-4}
    \cmidrule(r){5-8}
    \gc{Supervised} & \multirow{5}*{Cascade R-CNN} & \multirow{5}*{R-50-FPN} & \multirow{5}*{12e} & \gc{26.3} & \gc{19.7} & \gc{25.3} & \gc{30.3} \\
    CSD~\cite{jeong2019consistency} & ~ & ~ & ~ & 26.8 & 19.9 & 25.8 & 31.0 \\
    STAC~\cite{sohn2020simple} & ~ & ~ & ~ & 27.5 & 20.3 & 26.3 & 32.1 \\
    Unbiased Teacher~\cite{liu2021unbiased} & ~ & ~ & ~ & 28.6 & 20.8 & 27.9 & 32.8  \\
    Soft Teacher~\cite{xu2021end} & ~ & ~ & ~ & 29.2 & 21.1 & 28.4 & 33.7 \\
    LabelMatch~\cite{chen2022dense} & ~ & ~ & ~ & 29.4 & 20.3 & 29.2 & 33.8 \\
    \rowcolor{GrayBG}
    CascadeMatch (\emph{ours}) & ~ & ~ & ~ &  \bf{30.5}  & \bf{23.1} & \bf{29.7} & \bf{34.7} \\
    \cmidrule(r){1-1} 
    \cmidrule(r){2-4}
    \cmidrule(r){5-8}
    \gc{Supervised} & \multirow{3}*{Cascade R-CNN} & \multirow{3}*{R-101-FPN} & \multirow{3}*{12e} & \gc{27.1} & \gc{20.3} & \gc{26.1} & \gc{31.1} \\
    Unbiased Teacher~\cite{liu2021unbiased} & ~ & ~ & ~ & 31.0 & 24.6 & 30.2 & 35.0 \\
    \rowcolor{GrayBG}
    CascadeMatch (\emph{ours}) & ~ & ~ & ~ & \bf{32.9} & \bf{26.5} & \bf{31.8} & \bf{36.8} \\
    \cmidrule(r){1-1}
    \cmidrule(r){2-4}
    \cmidrule(r){5-8}
    \gc{Supervised} & \multirow{3}*{Sparse R-CNN} & \multirow{3}*{PVT} & \multirow{3}*{30e} & \gc{31.7} & \gc{23.5} & \gc{29.5} & \gc{38.0}  \\
    Unbiased Teacher~\cite{liu2021unbiased}  & ~ & ~ & ~ & 33.5 & 24.6 & 31.4 & 40.2 \\
    \rowcolor{GrayBG}
    CascadeMatch (\emph{ours}) & ~ & ~ & ~ & \bf{35.2} & \bf{27.5} & \bf{33.2} & \bf{41.1} \\
    \bottomrule
    \end{tabular}}
\end{table*}

\begin{table}[t]
\caption{Results on COCO-LT \textit{validation} set set.
The symbols \apone, \aptwo, \apthree and \apfour denote the bin of $[1, 20)$, $[20, 400)$, $[400, 8000)$, $[8000, -)$ training instances. The symbol `UT' is the abbreviation of the Unbiased Teacher~\cite{liu2021unbiased} algorithm.}
\label{tab:mainresults_cocolt}
\centering
\scalebox{1.0}{\tablestyle{3pt}{1.0}
\begin{tabular}{cx{70}x{20}x{20}x{20}x{20}}
\toprule
    Method  & AP & \apone & \aptwo & \apthree & \apfour \\
\cmidrule(r){1-1}
\cmidrule(r){2-6}
    \gc{Supervised} & \gc{25.4} & \gc{2.5} & \gc{16.2} & \gc{29.9} & \gc{33.7} \\
\cmidrule(r){1-1}
\cmidrule(r){2-6}
    CSD& 25.9 \td{+0.5} & 2.0 & 15.2 & 32.1 & 34.0 \\
    STAC & 26.4 \td{+1.0} & 2.2 & 16.3 & 32.4 & 34.1 \\
    UT   & 26.7 \td{+1.3} & 2.2 & 18.0 & 31.8 & 34.3 \\
\rowcolor{GrayBG}
    Ours &\bf{27.8} \td{+2.4} & \bf{4.0} & \bf{20.4} & \bf{32.4} & \bf{34.5} \\
\bottomrule
\end{tabular}}
\end{table}

\begin{table}[t]
    \centering
    \caption{Comparisons of training memory (MB), training time  $T_{\text{train}}$ (sec/iter) and inference time $T_{\text{test}}$ (sec/iter) on the LVIS dataset.}
    \label{tab:computation_budgets}
    \tablestyle{2pt}{1.0}
    \begin{tabular}{cx{40}x{40}x{40}}
    \toprule
    Method & Memory & $\operatorname{T}_{\mathrm{train}}$ & $\operatorname{T}_{\mathrm{test}}$ \\
    \cmidrule(r){1-1}
    \cmidrule(r){2-2}
    \cmidrule(r){3-3}
    \cmidrule(r){4-4}
    \gc{Supervised} & \textbf{5889} & \textbf{0.2248} & \textbf{0.2694} \\
    CSD & 6452 & 0.3310 & 0.2767 \\
    STAC & 6801 & 0.4110 & 0.2702 \\
    Unbiased Teacher & 7366 & 0.4616 & 0.2761 \\
    Soft Teacher & 8029 & 0.4589 & 0.2718 \\
    LabelMatch & 8240 & 0.4918 & 0.2698 \\
    \rowcolor{GrayBG}
    Ours & 7432 & 0.4733 & 0.2734 \\
    \bottomrule
    \end{tabular}
\end{table}

\begin{figure*}[t!]
    \includegraphics[width=1.0\textwidth]{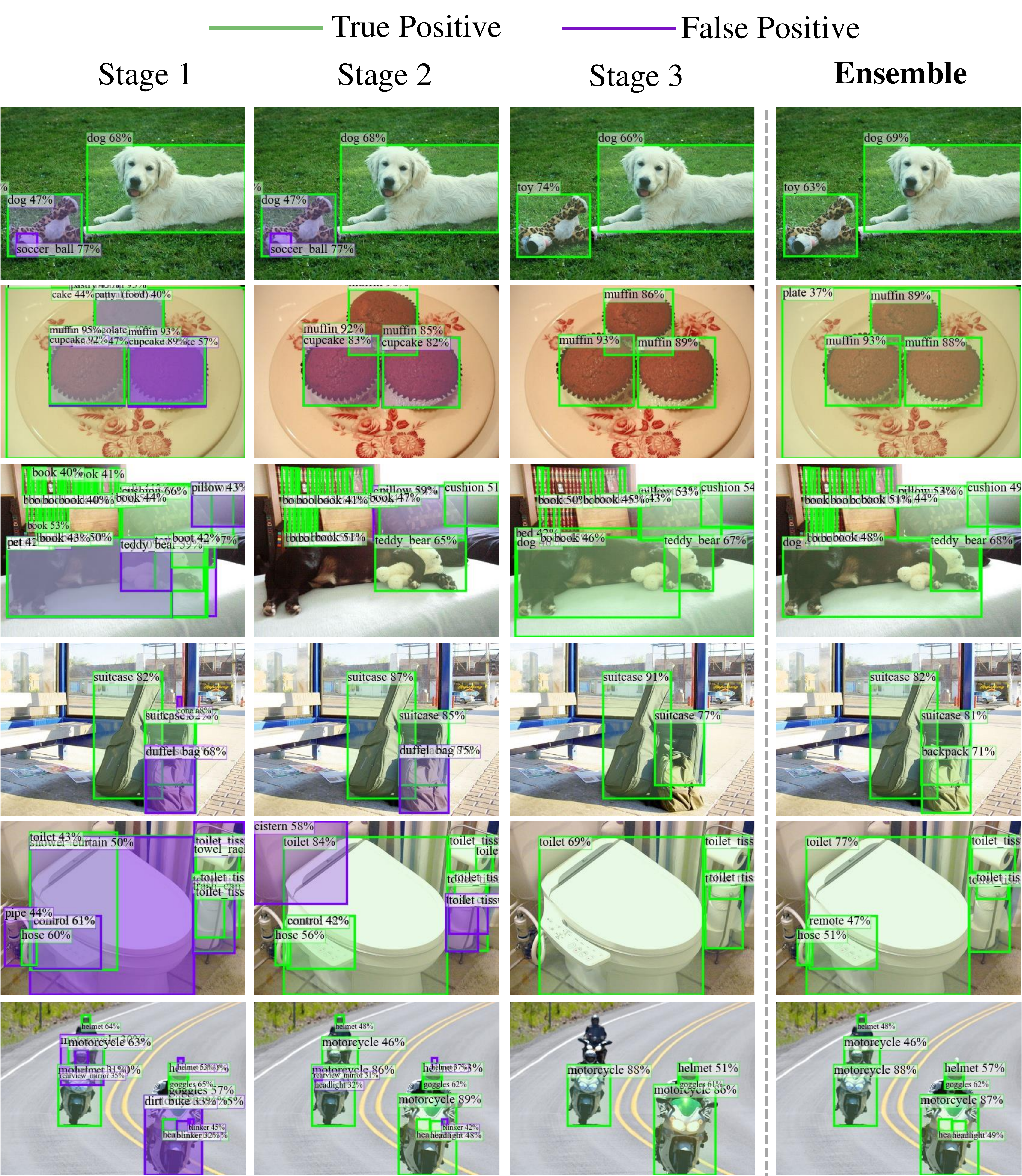}
    \centering
    \caption{
    The pseudo labels generated on the LVIS \textit{training} dataset under the \textbf{semi-supervised object detection setting~(SSOD)} setting.
    The \textcolor{Green}{\bf{green}} color refers to the true-positive predicted results; \textcolor{Purple}{\bf{purple}} color refers to false-positive detection results (Zoom in for best view).
    }
    \label{fig:vis_pseudo}
\end{figure*}

\smallsec{Baselines}
In this section, we compare our method against the supervised baseline (without using the unlabeled data) and state-of-the-art semi-supervised learning methods on the LVIS v1.0 and COCO-LT datasets.
We select four representative semi-supervised detection algorithms to compare with: 1) CSD~\cite{jeong2019consistency} is a consistency regularization-based algorithm that forces the detector to make identical predictions under different augmentations.
2) STAC~\cite{sohn2020simple} is a pseudo-labeling-based method that uses an off-line supervised model as a teacher to extract pseudo-labels.
3) Unbiased Teacher~\cite{liu2021unbiased} and 4) Soft Teacher~\cite{xu2021end} are also a pseudo-labeling-based method that uses the exponential moving average~(EMA) ensemble to provide a strong teacher model. Soft Teacher uses extra box jittering augmentation to further boost the performance. 5) LabelMatch~\cite{chen2022label} introduces a re-distribution mean teacher based on the KL divergence distribution between teacher and student models. Unbiased Teacher, Soft Teacher and LabelMatch are strong baselines so the comparison with them can well demonstrate the effectiveness of our approach.
We use the open-source code provided by the authors and re-train the model on the LVIS v1.0 and COCO-LT datasets, respectively. All baselines and our approach use the Equalization Loss v2 (EQL v2)~\cite{tan2020equalization2} as the default classification loss. EQL v2 improves the model's recognition ability by down-weighting negative gradients for rare classes.

\smallsec{Results on LVIS v1.0}
Table~\ref{tab:mainresults_lvis} shows the results on LVIS. When using Cascade R-CNN and ResNet50 as the backbone, our approach improves \fixedap from the supervised baseline’s $26.3$ to $30.5$, achieving $4.2$ mAP improvement. Compared with LabelMatch, which is the strongest baseline, CascadeMatch still maintains clear advantages. Overall, the results presented in the experiments validate the effectiveness of the cascade pseudo-labeling design and the adaptive pseudo-label mining mechanism.

\smallsec{Results on COCO-LT}
As shown in Table~\ref{tab:mainresults_cocolt},  an absolute improvement of $2.4$ in mAP is obtained by CascadeMatch over the supervised baseline on COCO-LT. The results indicates the generalizability of the CascadeMatch across multiple datasets.

\smallsec{Large Model $\&$ More Architectures}
Table~\ref{tab:mainresults_lvis} also shows the results using other architectures. With ResNet101 as the backbone under the Cascade R-CNN framework, CascadeMatch outperforms Unbiased Teacher by $1.9$ \fixedapr and $1.6$ \fixedapc. With Sparse R-CNN and the Transformer encoder, CascadeMatch also gains clear improvements: $1.7$ \fixedap and $2.9$ \fixedapr.
Such results show that our proposed method is general to various architectures.

\smallsec{Computation Budgets}
We report the training memory, training time, and inference time against the supervised baseline and different semi-supervised methods, as shown in Table~\ref{tab:computation_budgets}.
All the methods are based on the Cascade-RCNN framework with the ResNet50-FPN backbone and report on one Nvidia V100 GPU.
We can see that when compared with the supervised baseline, CSD has an increased memory footprint and training time because of the extra steps during training like data augmentation and forward pass on unlabeled data. For pseudo-labeling methods, like Unbiased Teacher and LabelMatch, the training cost further increases with the generation of pseudo-labels. Our CascadeMatch method shares similar memory and training time as Unbiased Teacher, thus is comparable to recent semi-supervised methods in terms of the training cost. We also find all these methods (including ours) have negligible overhead in the inference stage, with almost the same inference time as the supervised learning baseline.

\smallsec{Qualitative Results}
We show some pseudo-labeling visualization results under the semi-supervised object detection~(SSOD) setting in Figure~\ref{fig:vis_pseudo}.
Since we set a progressive confidence threshold $\tau$ from stage 1 to 3, we observe that stage 1 focuses on generating redundant pseudo labels with high recall and some false positive results (in \textcolor{Purple}{\bf{purple}}).
In contrast, stage 3 prefers high precision pseudo labels, but some prediction results may be missed.
The ensemble of pseudo label predictions is of high quality and controls the precision-recall trade-off well.
According to the quantitative results in Table~\ref{tab:mainresults_lvis} and the qualitative results shown in Figure~\ref{fig:vis_pseudo}, we can conclude that CascadeMatch benefits from more accurate pseudo-labels it estimates for the unlabeled data.

\subsection{Sparsely Annotated Object Detection}

\begin{table}[t]
\tablestyle{7pt}{1.0}
	\caption{
	Experiment results under the Sparsely annotated object detection (SAOD) setting where missing labels exist in the training set.
    We follow previous studies~\cite{zhang2020solving,wang2021co} to build a modified LVIS dataset where we randomly erase the annotations by 20\% and 40\% per object category.}
    \label{tab:saod}
	\centering
    \begin{tabular}{cccccc}
    \toprule
    Missing Ratio & Ours & AP & \apr & \apc & \apf \\
    \cmidrule(r){1-2}
    \cmidrule(r){3-6}
    \multirow{2}*{40\%}  & \gxmark & \gc{22.5} & \gc{10.4} & \gc{20.9} & \gc{29.6} \\
    ~                    & \cmark & \GrayBG{\textbf{24.2}} & \GrayBG{\textbf{13.7}} & \GrayBG{\textbf{22.4}} & \GrayBG{\textbf{30.9}} \\
    \cmidrule(r){1-2}
    \cmidrule(r){3-6}
    \multirow{2}*{20\%}  & \gxmark   & \gc{24.7} & \gc{14.3} & \gc{22.7} & \gc{31.4} \\
    ~                    & \cmark    & \GrayBG{\textbf{26.7}} & \GrayBG{\textbf{17.2}} & \GrayBG{\textbf{25.1}} & \GrayBG{\textbf{32.8}} \\
    \bottomrule
\end{tabular}
\end{table}

\smallsec{Background}
The standard semi-supervised learning setting in object detection assumes that training images are fully annotated. A more realistic setting that has received increasing attention from the community is sparsely annotated object detection~\cite{wu2018soft,zhang2020solving,wang2021co,zhou2021probabilistic}, or SAOD. In the previous experiments, we have shown that CascadeMatch performs favorably against the baselines with clear improvements. In this section, we unveil how CascadeMatch fares under the SAOD setting.

In SAOD, some images are only partially annotated, meaning that not all instances in an image are identified by bounding boxes. Such a phenomenon is in fact common in existing large-vocabulary datasets like the previously used LVIS~\cite{gupta2019lvis} dataset. Unidentified instances are simply treated as background in existing semi-supervised approaches. As a consequence, no supervision will be given to the model with respect to those instances. Different from SSOD, the goal in SAOD is to identify instances with missing labels from the training set.

\smallsec{Experimental Setup}
We use LVIS as the benchmark dataset. CascadeMatch is compared with Federated Loss~\cite{zhou2021probabilistic}, which serves as a strong baseline in this setting. Concretely, Federated Loss ignores losses of potentially missing categories and thus uses only a subset of classes for training. To facilitate evaluation, we follow previous studies~\cite{zhang2020solving,wang2021co} to build a modified LVIS dataset where a certain percentage of annotations within each category are randomly erased. We choose the 20\% and 40\% as the percentage numbers. The baseline model is the combination of Cascade R-CNN~\cite{cai2019cascade} and Federated Loss.
Noted that it is common to select 50\% erasing ratio~\cite{zhang2020solving,wang2021co} for balanced datasets. However, for long-tailed datasets erasing 50\% annotations would lead to significantly fewer annotations for rare classes (23.73\% of rare classes will have zero annotations).  We chose the 20\% and 40\% ratios to cover different scenarios (95.54\% and 88.76\% of rare classes are preserved that have at least one annotation).

\smallsec{Results}
We experimented with the 20\% and 40\% missing ratios on our modified LVIS dataset. The results are reported in Table~\ref{tab:saod} where the checkmark symbol means that CascadeMatch is applied to the model. In both settings, we observe a clear margin between CascadeMatch and the baseline: +1.8\% and +2.0\% gains in terms of overall AP under the settings of 20\% and 40\% missing ratios, respectively. Notably, the gains are more apparent for the rare classes, with +3.3\% and +2.9\% gains for the two settings, respectively. The quantitative results shown in Table~\ref{tab:saod} strongly demonstrate the ability of CascadeMatch in dealing with the SAOD problem.

\smallsec{Qualitative Results}
\begin{figure*}[t!]
    \centering
    \includegraphics[width=0.98\textwidth]{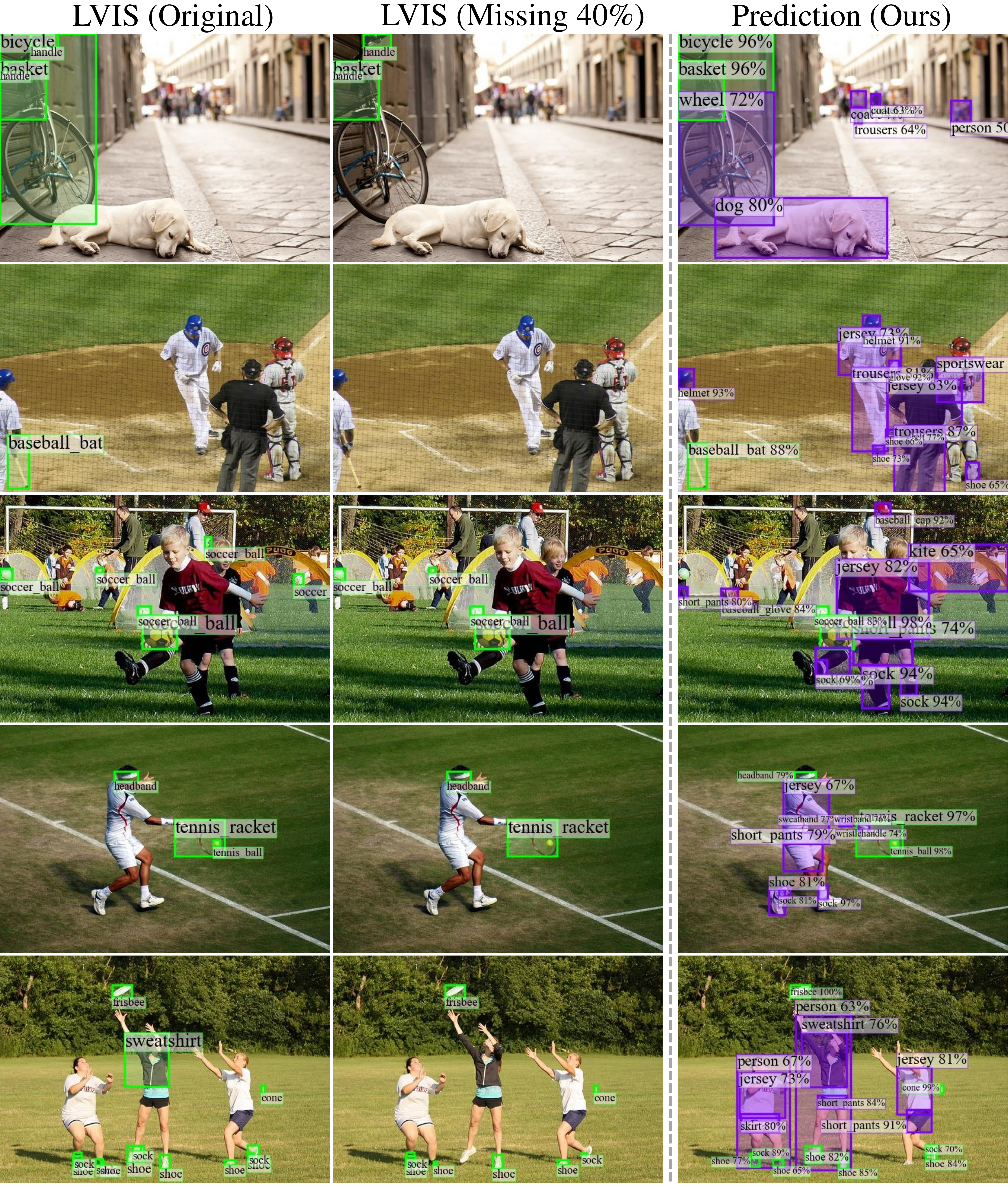}
    \caption{The pseudo labels generated on the LVIS \textit{training} dataset under the \textbf{sparsely-annotated object detection setting~(SAOD)} setting.
    In the third column, \textcolor{Green}{\bf{green}} color refers to the predicted results that can be found in the ground truth of the first column; \textcolor{Purple}{\bf{purple}} color refers to predicted results that are also missing in the original LVIS dataset (Zoom in for best view).
    }
    \label{fig:vis_saod}
\end{figure*}
We also show the visualization results of the pseudo-labeling under the sparsely-annotated object detection~(SAOD) setting in Figure~\ref{fig:vis_saod}.
The first column refers to the ground truth labels from the original LVIS dataset.
The second column shows our modified sparsely-annotated LVIS dataset where some annotations are randomly removed with a 40\% missing rate and serves as the training set under the SAOD setting.
The third column contains the prediction results of CascadeMatch.
We observe that CascadeMatch can recover some labels.
Since the original LVIS datasets is sparsely-annotated, CascadeMatch can also detect objects whose labels are missing in the original LVIS dataset. The qualitative results in Figure~\ref{fig:vis_saod} explain the excellent performance of CascadeMatch on the SAOD task.

\section{Limitation}
\label{sec:limitation}
The trade-off between speed and performance is one of the key research problems in the area of object detection~\cite{liu2016ssd,ren2015faster,lin2017focal,cai2019cascade,tian2019fcos,carion2020end}. It has been widely acknowledged that achieving a perfect speed-performance trade-off is extremely difficult~\cite{huang2017speed}. To obtain a high-performance detector, one has to sacrifice on the speed, and vice versa. In this work, our CascadeMatch processes data in a cascade manner, which leads to longer training time and slower inference speed compared to the single-stage detector counterpart. However, given that the majority of computation takes place in the backbone while the detection heads are generally ``lightweight'' (as they only consist of a few fully connected layers), the lower speed is outweighed by the improvements in performance. To further improve the efficiency in real-world deployment, one could apply model compression techniques to reduce the model size, and design more lightweight architectures for the cascade detection heads.

\section{Conclusion}
Our research addresses an important but largely under-studied problem in object detection, concerning both long-tailed data distributions and semi-supervised learning. The proposed approach, CascadeMatch, carefully integrates pseudo-labeling, coupled with a cascade design and an adaptive threshold tuning mechanism, into a variety of backbones and detection frameworks, such as the widely used region proposal-based detectors and more recent fully end-to-end detectors. The results strongly demonstrate that CascadeMatch is a better design than existing state-of-the-art semi-supervised detectors in handling long-tailed datasets such as LVIS and COCO-LT. The capability to cope with the sparsely-annotated object detection problem is also well justified.

\smallsec{Data Availability Statements} The datasets analysed during this study are all publicly available for the research purpose - the \href{https://www.lvisdataset.org/}{LVIS} and \href{https://cocodataset.org/#home}{COCO} datasets.

\smallsec{Acknowledgement} This study is supported under the RIE2020 Industry Alignment Fund Industry Collaboration Projects (IAF-ICP) Funding Initiative, as well as cash and in-kind contribution from the industry partner(s). It is also partly supported by the NTU NAP grant and Singapore MOE AcRF Tier 2 (MOE-T2EP20120-0001).


%
%


\printbibliography

\end{document}